\documentclass[lettersize,journal]{IEEEtran}
\usepackage{amsmath,amsfonts}
\usepackage{booktabs}
\usepackage{algorithmic}
\usepackage{algorithm}
\usepackage{array}
\usepackage[caption=false,font=normalsize,labelfont=sf,textfont=sf]{subfig}
\usepackage{textcomp}
\usepackage{stfloats}
\usepackage{url}
\usepackage{verbatim}
\usepackage{graphicx}
\usepackage{cite}
\usepackage[colorlinks=true, linkcolor=blue, citecolor=blue, urlcolor=blue]{hyperref}

\usepackage[table]{xcolor}
\usepackage{colortbl}
\usepackage{multirow}
\usepackage{multicol}
\usepackage{amssymb}
\usepackage{hyperref}

\newcommand{\et}{\textit{et al.}}
\begin{document}

\title{Complementary and Contrastive Learning for Audio-Visual Segmentation}

\author{Sitong Gong, Yunzhi Zhuge, Lu Zhang, Pingping Zhang, Huchuan Lu,~\IEEEmembership{Fellow~IEEE}
\thanks{Sitong Gong, Yunzhi Zhuge and Lu Zhang are with the School of Information and Communication Engineering, Dalian University of Technology, Dalian 116081, China. (e-mail: stgong@mail.dlut.edu.cn; zgyz@dlut.edu.cn; zhangluu@dlut.edu.cn)}
\thanks{Pingping Zhang and Huchuan Lu are with the School of Future Technology and the School of Artificial Intelligence, Dalian University of Technology, Dalian 116081, China. (email: zhpp@dlut.edu.cn; lhchuan@dlut.edu.cn)}}


\markboth{IEEE Transactions on Multimedia}%
{Shell \MakeLowercase{\textit{et al.}}: A Sample Article Using IEEEtran.cls for IEEE Journals}


\maketitle

\begin{abstract}
Audio-Visual Segmentation (AVS) aims to generate pixel-wise segmentation maps that correlate with the auditory signals of objects. 
This field has seen significant progress with numerous CNN and Transformer-based methods enhancing the segmentation accuracy and robustness. Traditional CNN approaches manage audio-visual interactions through basic operations like padding and multiplications but are restricted by CNNs' limited local receptive field.
More recently, Transformer-based methods treat auditory cues as queries, utilizing attention mechanisms to enhance audio-visual cooperation within frames. Nevertheless, they typically struggle to extract multimodal coefficients and temporal dynamics adequately. To overcome these limitations, we present the Complementary and Contrastive Transformer (CCFormer), a novel framework adept at processing both local and global information and capturing spatial-temporal context comprehensively.  Our CCFormer initiates with the Early Integration Module (EIM) that employs a parallel bilateral architecture, merging multi-scale visual features with audio data to boost cross-modal complementarity. To extract the intra-frame spatial features and facilitate the perception of temporal coherence, we introduce the Multi-query Transformer Module (MTM), which dynamically endows audio queries with learning capabilities and models the frame and video-level relations simultaneously. Furthermore, we propose the Bi-modal Contrastive Learning (BCL) to promote the alignment across both modalities in the unified feature space.
Through the effective combination of those designs, our method sets new state-of-the-art benchmarks across the S4, MS3 and AVSS datasets.
Our source code and model weights will be made publicly available at \href{https://github.com/SitongGong/AVS_CCFormer}{https://github.com/SitongGong/CCFormer}.
\end{abstract}

\begin{IEEEkeywords}
Audio-Visual Segmentation, Complementary Learning, Multimodal Transformer, Contrastive Learning.
\end{IEEEkeywords}

\section{Introduction}
\IEEEPARstart{R}ECENTLY, the capacity for machines to coherently merge auditory and visual cues is becoming increasingly crucial. An emerging field of this evolution is audio-visual segmentation (AVS)~\cite{zhou2022audio}, which involves segmenting video frames to identify objects corresponding to specific audio clips.
Unlike previous audio-visual tasks such as speech enhancement~\cite{wang2020robust,xiong2022look}, sound source separation~\cite{he2021modeling,zhu2022visually}, and audio-visual event localization~\cite{xue2021audio,jiang2023leveraging,feng2023css}, AVS tackles the intricate challenge of interpreting the often elusive and dynamic aspects of sound to precisely identify and segment objects that are audibly discernible. 
The introduction of AVS has enhanced the capability of Large Multimodal Models (LMMs)~\cite{he2024multi} to perceive audio semantic information at a fine-grained level. This advancement has significant potential for applications in advancing fields, such as embodied intelligence~\cite{li2024embodied}.

\begin{figure*}
    \centering
    \includegraphics[width=1\linewidth]{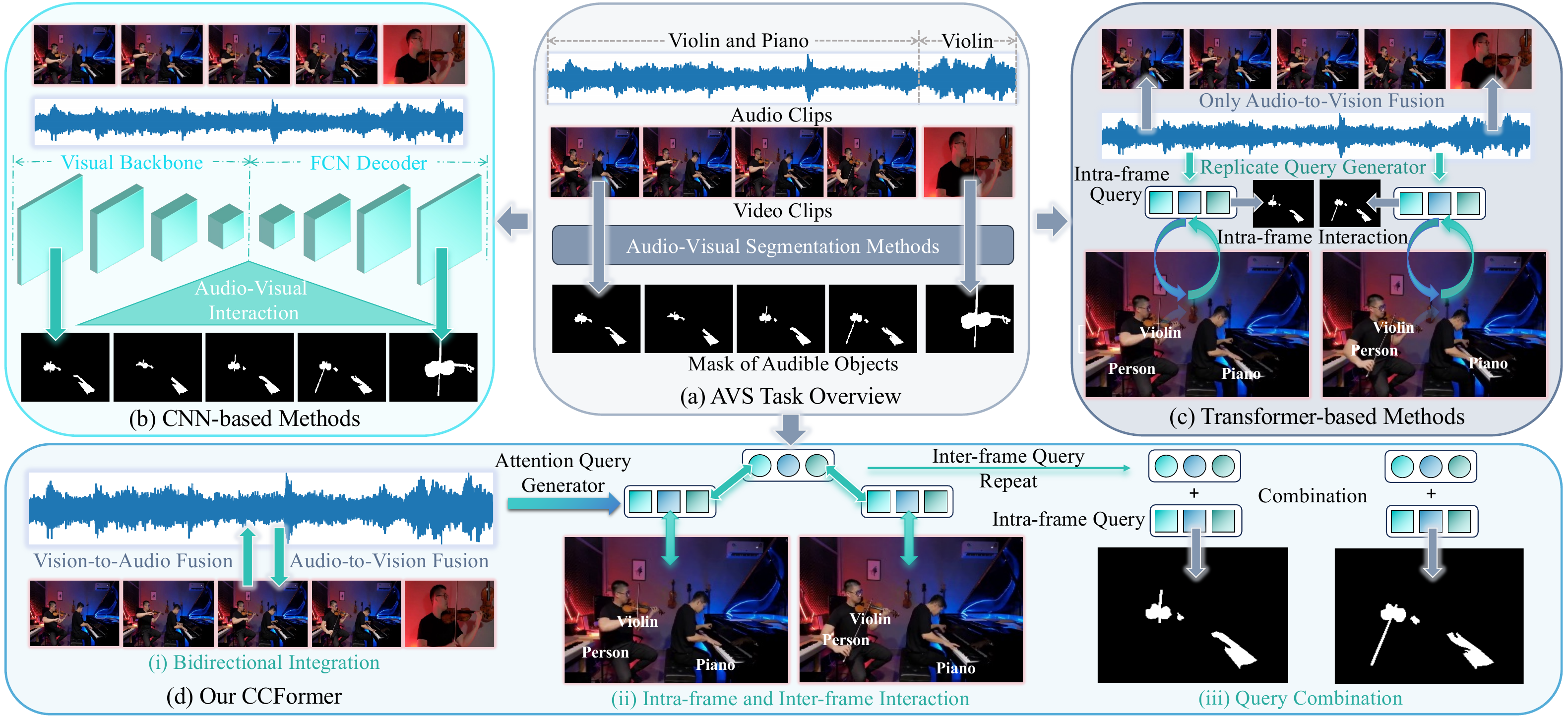}
    \caption{Comparative frameworks in audio-visual segmentation (AVS): (a) AVS aims to produce mask sequences according to audio prompts. Previous approaches mainly rely on (b) FCN or (c) Transformer blocks to perform per-frame audio-visual integration, insufficiently addressing the intrinsic temporal coherence in both video and audio sources. On the contrary, (d) our CCFormer realizes a comprehensive spatial-temporal interaction between two modalities by building bidirectional integration, intra- and inter-frame interaction and query combination.
    }
    \vspace{-1mm}
    \label{fig:methods comparison}
\end{figure*}

Existing AVS methods mainly fall into two paradigms: FCN-based models~\cite{zhou2022audio,hao2023improving}, exemplified in Fig.~\ref{fig:methods comparison}(b), and Transformer-based models~\cite{ling2023hear,gao2024avsegformer,huang2023discovering}, illustrated in Fig.~\ref{fig:methods comparison}(c). To strengthen the guidance of audio sequence, FCN-based methods perform dense audio-visual interaction among CNN features. However, the restricted receptive fields of CNNs limit their capability to capture the global spatio-temporal coherence of audio-visual data,  resulting in inaccurate localization and incomplete segmentations of the sounding objects. On the contrary, Transformer-based approaches rely on multi-head attention for dense audio-visual integration and learnable queries to capture object concepts comprehensively, leading to significant improvement against most FCN-based methods.  
Although the adoption of Transformer architectures facilitates enhanced long-range and comprehensive multimodal integration, existing methods still exhibit inherent limitations:
\begin{itemize}
    \item Most Transformer-based methods, such as 
    AVSegFormer~\cite{gao2024avsegformer} and AQFormer~\cite{huang2023discovering}, predominantly employ the attention mechanism for either unidirectional feature integration or multi-scale visual enhancement at initial processing stages.  These methods lead to imbalanced information fusion across modalities, compelling the model to overly rely on single-modal features. Such reliance results in the accumulation of redundant information and diminishes the decoder's ability to detect the salient audible regions. Consequently, the inadequate cross-modal interaction prevents the effective utilization of the temporal coherence inherent in audio-visual data, ultimately compromising the accuracy of identifying and segmenting sound-producing objects.
    \item In earlier approaches~\cite{gao2024avsegformer,li2023catr}, audio queries are integrated with visual features frame-by-frame. This integration strategy limits the model’s ability to capture comprehensive spatial correlations at the frame level and temporal correlations at the video level, both of which are essential for accurate object localization and the generation of consistent mask sequences.
The imbalance in capturing spatio-temporal information would limit the model's capability to effectively model global features and establish fine-grained perception of the visually relevant regions of sound sources.
\end{itemize}


To cope with the above limitations, we propose CCFormer, a novel framework designed to achieve complementary and contrastive learning through intra-frame and inter-frame interactions for audio-visual segmentation. 
Firstly, we introduce an Early Integration Module (EIM) to establish a rich complementary interdependence between audio and visual modalities. This module facilitates concurrent audio-to-vision and vision-to-audio interactions, enriching each modality with information derived from the other. Secondly, our Multi-query Transformer Module (MTM) enhances object identification across frames. It begins with intra-frame queries generated by the Attention Query Generator, leveraging attention mechanisms for interactions with enhanced audio features. These queries then decode sounding object information by attending to corresponding multi-scale visual features. Additionally, we reconfigure these queries to enable spatial-temporal interaction and integrate global information across frames. Finally, we implement Bi-modal Contrastive Learning (BCL) to bolster our model's perceptual learning across modalities. Here, enhanced and original audio features are mapped into a multimodal feature space, with a contrastive loss employed for modality alignment before processing through the Transformer.

Our contributions can be summarized as follows:
\begin{itemize}
\item We introduce CCFormer, a novel Complementary and Contrastive Transformer, specifically designed for audio-visual segmentation. 
\item We carefully design multiple modules to facilitate complementary and contrastive cross-modal learning. To begin with, the EIM is employed to promote bidirectional multimodal fusion. Then, within the MTM, the intra-frame and inter-frame queries construct the explicit association to capture spatio-temporal features. We further introduce the BCL module to strengthen the representation similarity between audio and visual modalities.
\item Experimental evaluations demonstrate that CCFormer consistently achieves state-of-the-art performance on both AVSBench-object and AVSBench-semantic datasets, surpassing previous methods by a considerable margin.
\end{itemize}

\section{Related Works}
\subsection{Sound Source Localization}
Sound source localization (SSL) aims at pinpointing the sound sources within video frames under the guidance of the provided audio signals. Qian \et~\cite{qian2020multiple} employed class activation mapping to extract class-specific representations, implementing multi-source localization by performing cross-modal alignment in a coarse-to-fine manner.  Hu \et~\cite{hu2022mix} created the graph for the image and predicted sounds and executed random walk to learn a similarity metric, maximizing the probability of cycle-consistency for the walk. 
AGVN~\cite{mo2023audio} introduced a category-aware audio-visual grouping strategy to separate the semantic characteristics of each sound source from the mixture, learning the high-level semantic information. 
SLAVC~\cite{mo2022closer} applied a momentum encoder and cross-modal contrastive learning to achieve weakly-supervised audio-visual source localization. Existing SSL methods are only capable of perceiving the general location of sounding objects, lacking the precision to perform pixel-level classification of prominent sound sources, thereby restricting the models' recognition performance and applicability. Consequently, the task of audio-visual segmentation has been introduced~\cite{zhou2022audio}, which aims to endow models with the capability to locate and segment complex sound sources accurately.

\subsection{Audio-Visual Segmentation.}
The audio-visual segmentation (AVS) task was first introduced by Zhou \textit{et al.}~\cite{zhou2022audio}, aiming to perform pixel-level segmentation of sound sources, improving downstream applications, such as video understanding~\cite{jelodar2018long} and augmented reality~\cite{zhang2018inverse}. 
They also proposed a baseline method employing ASPP and TPAVI modules for multimodal fusion.
Building on this foundation, AVSBiGen~\cite{hao2023improving} introduced the vision-to-audio projection for audio reconstruction while employing the Visual Correlation Attention Module to enhance the association between adjacent frames. 
CNN-based methods like TPAVI, however, are limited by the narrow receptive fields and static nature of 2D convolution. In contrast, Transformer-based approaches~\cite{ling2023hear,gao2024avsegformer,liu2023audio} demonstrated improved performance in feature extraction and interaction. For instance, 
BAVS~\cite{liu2024bavs} focused on eliminating background noises and establishing audio-visual correspondences by incorporating multimodal foundation knowledge. 
AVSegFormer~\cite{gao2024avsegformer} employed the Transformer decoder to facilitate the association between audio features and sounding objects from corresponding frames. AQFormer~\cite{huang2023discovering} linked object queries to sounding objects and introduced the ABTI Module for temporal modeling, while CATR~\cite{li2023catr} employed various modules to enhance multimodal feature fusion. AVSAC~\cite{chen2024bootstrapping} utilized the continuous bidirectional fusion approach to strengthen the auditory cues and maintain the balance of cross-modal interactions. 
Our method distinguishes itself from these prior works by integrating inter-frame cross-modal feature interaction with contrastive learning to attain superior performance.

\subsection{Transformer for Video Segmentation}
Video segmentation~\cite{zhou2022survey}, as surveyed by Zhou \et~\cite{zhou2022survey}, involves dividing video frames into multiple segments that correspond to distinct objects, tailored to specific application requirements. This field includes several sub-disciplines such as video object segmentation (VOS), which isolates particular objects throughout a video sequence; referring video object segmentation (RVOS), which segments objects based on a descriptive query; and video instance segmentation (VIS), which identifies and segments individual object instances across frames. Transformer-based methods~\cite{wang2021end,hwang2021video} have become predominant as they excel in handling long-term sequences and occlusions. VisTR~\cite{wang2021end} utilized the DETR~\cite{carion2020end} architecture for video instance segmentation. Subsequent models like IFC~\cite{hwang2021video} and VITA~\cite{heo2022vita} enhanced inter-frame interactions through memory tokens and object queries. In video object segmentation~\cite{luo2017video,sun2022starting,chen2019multilevel}, AOT~\cite{yang2021associating} employed a long short-term Transformer for multi-object association, leveraging hierarchical matching and propagation based on initial one-shot masks.
As for referring video object segmentation~\cite{wu2022language}, the goal is to segment the target object given its corresponding language description. 
SOC~\cite{luo2023soc} adopted the object cluster module for video-level information collection and a contrastive loss for bi-modal alignments. Similarly, MUTR~\cite{yan2023referred} built upon the architecture of ReferFormer~\cite{wu2022language} by integrating the temporal fusion module, facilitating text features to focus on inter-frame visual features. 
BIFIT~\cite{lan2023bidirectional} enhanced the Transformer decoder by incrementally inserting an inter-frame interaction module after each layer, learning the spatio-temporal features of the referred objects. 
Inspired by these endeavors, we leverage Transformers to model intra-frame and inter-frame relationships, aiming for improved accuracy in audio-visual segmentation.

\subsection{Contrastive Learning in Audio-Visual Tasks}
Contrastive learning, which aims at learning more representative features through the contrasting of positive and negative pairs, has received remarkable attention in recent years~\cite{chen2020simple,khosla2020supervised,tian2020makes}. 
Within the scope of audio-visual perceptions, EPCFormer~\cite{chen2023epcformer} proposed the expression contrastive loss to gather the audio and linguistic features with the same semantic content in the representation space. 
Reference~\cite{senocak2023sound} proposed improvements in sound source localization by emphasizing semantic consistency and enhancing the spatial similarity of cross-modal features. This approach utilizes advanced localization techniques alongside instance-level similarity functions, which together serve to refine the alignment of audio and visual data. 
WS-AVS~\cite{mo2023weakly} introduced a multi-scale, multi-instance contrastive learning scheme, using maximum pooling and similarity calculation across various visual scales and audio features within the same batch, facilitating the model focusing on the visual regions most relevant to the audio. 
Diff-AVS~\cite{mao2023contrastive} integrated the latent diffusion paradigm with the principles of contrastive representation learning to elucidate the audio-visual correspondence. 
Differing from previous methods, our contribution encompasses the development of a Bi-modal Contrastive Loss specifically engineered for the refinement of audio-visual segmentation.  The innovative loss function significantly refines the alignment between audio and visual modalities, promoting the proximity of audio to its corresponding visual modalities.

\section{Method}
\subsection{Overall Framework}
\begin{figure*}
    \centering
    \includegraphics[width=1\linewidth]{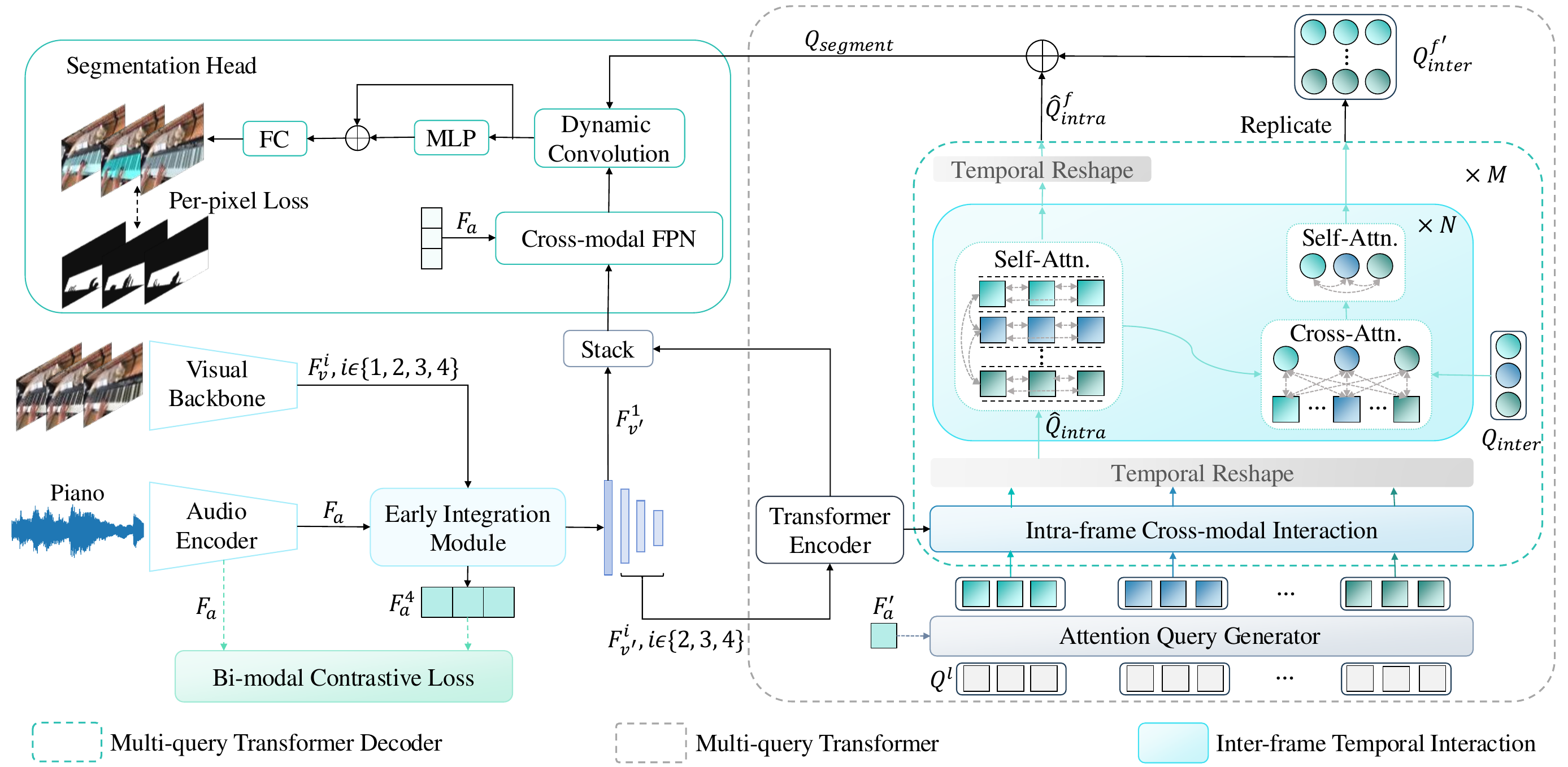}
    \caption{Overview of the CCFormer architecture. 
    The model begins by employing the Early Integration Module to perform cross-frame bidirectional fusion of the audio and multi-scale visual features. 
    Subsequently, the Multi-query Transformer Module introduces two types of queries and a progressive interaction strategy. 
    Specifically, we utilize the Attention Query Generator to yield the intra-frame queries for cross-modal interactions with visual features within a single frame. Then the inter-frame queries are initialized to perform temporal interactions with the reshaped intra-frame queries. 
    After multiple iterations, both types of queries are combined to generate the mask embedding.
    Finally, Bi-modal Contrastive Loss calculates the contrastive loss between audio features before and after early integration, further aligning the multimodal features.
    \vspace{-1mm}
    }
\label{fig:2}
\end{figure*}

Given an audio sequence, our CCFormer aims to precisely locate and segment the sounding objects within a video sequence, mathematically formulated as $\mathcal{M}=\textit{CCFormer}(\mathcal{V},\mathcal{A})$. 
In this study, we denote the input video frames, audio signals, and generated segmentation masks by $(\mathcal{V},\mathcal{A},\mathcal{M}) = \{V_t, A_t, M_t\}_{t=1}^T$, respectively. Here, $T$ represents the length of the sequence. The overall framework of CCFormer is illustrated in Fig.~\ref{fig:2}. 
We first utilize visual and audio encoders to extract features for input video and audio sequences, which are denoted as $F_a$ and $F_v$. Then, we apply an Early Integration Module to fuse the multimodal features. The fused features are aligned with the initial audio queries and subsequently fed into the Multi-query Transformer Module (MTM) incorporating a Transformer Encoder, an Attention Query Generator, and a Multi-query Transformer Decoder (MTD). 
Specifically, the visual features output by EIM are first input into the Transformer Encoder for further enhancement while the initialized queries interact with enhanced audio features through the Attention Query Generator to generate intra-frame queries. Subsequently, the MTD takes the intra-frame queries and multi-scale visual features as input and performs feature interactions. 
In MTD, an Intra-frame Cross-modal Interaction (ICI) and an Inter-frame Temporal Interaction (ITI) collaborate to learn the spatio-temporal interaction of audio-visual features. Finally, we use a segmentation head to produce the mask sequence of the sounding objects. During the training process, in addition to the standard supervision applied to the outputs from the segmentation head, we propose a bi-modal contrastive learning strategy that is implemented subsequent to the early integration module, significantly enhancing the model's capability to perceive and integrate multimodal features. 


\subsection{Early Integration Module} 
\begin{figure}
    \centering
     \vspace{-2mm}
    \includegraphics[width=1\linewidth]{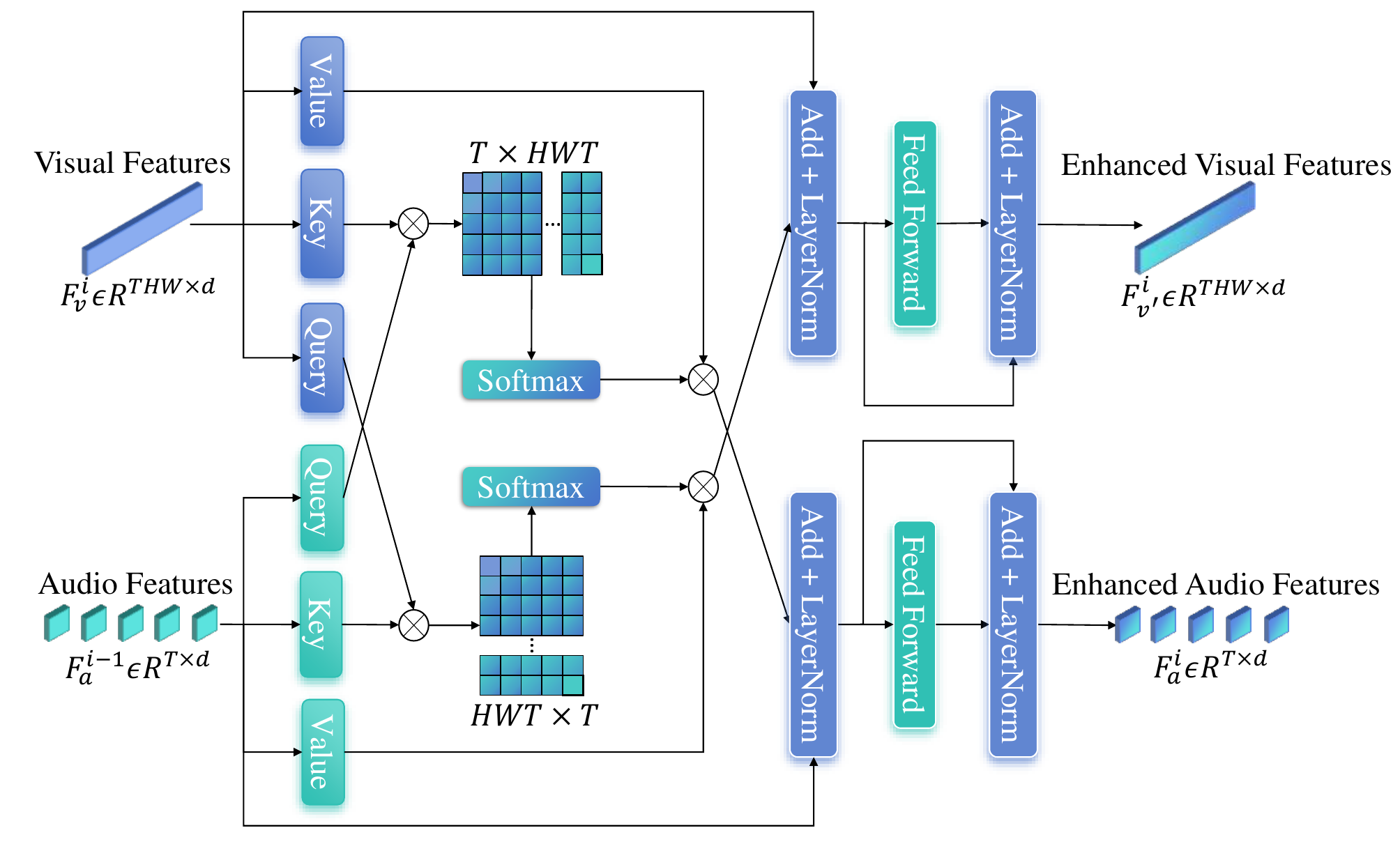}
    \vspace{-4mm}
    \caption{Illustration of our Early Integration Module. 
    It takes the integration process of $F^{i}_{v}$ and $F^{i-1}_{a}$ as an example. This module employs a bidirectional structure for initial feature fusion, where Audio-guided Vision and Vision-guided Audio Enhancement Modules leverage cross-attention mechanisms for bi-modal feature interaction and dimension alignment.}
    \vspace{-1mm}
    \label{fig:3}
\end{figure}

Previous methods primarily integrated audio-visual information on a per-frame basis, which often fails to adequately capture the temporal dynamics inherent in video and audio streams.
To address this limitation, we introduce an Early Integration Module (EIM) that optimizes dense multimodal feature fusion. The EIM utilizes a parallel bilateral architecture, consisting of an Audio-guided Vision Enhancement (AVE) Module and a Vision-guided Audio Enhancement (VAE) Module.
Detailed illustrations of the EIM are provided in Fig.~\ref{fig:3}. 

We begin by projecting both visual and audio features into the same dimension, $d$. Then we utilize the AVE Module and VAE Module to establish a multi-scale iterative integration framework. Specifically, for the visual features $F_v^i$ at the $i$-th scale, we initially apply AVE for scale-wise feature updating. This process can be represented as:   
\begin{equation}
F^{i}_{a\to v}=LN(Softmax(\frac{Q^{i}_{v}K^{\top}_{a}}{\sqrt{d_{head}}})V_{a}+F^{i}_{v}),
\end{equation}
\begin{equation} 
F^{i}_{v'}=LN(FFN(F^{i}_{a\to v})+F^{i}_{a\to v}),
\end{equation}
where $FFN$ represents the feed-forward network, $LN$ represents the layer normalization~\cite{ba2016layer}, $d_{head}$ represents the number of attention heads,  $F^{i}_{a\to v}$  represents the intermediate variable, and $F^{i}_{v}$ and $F^{i}_{v'}$ represent the original and enhanced visual features for the $i$-th scale. $Q^{i}_{v}$ represents the query projected by $F^{i}_{v}$ and $K_{a}$, $V_{a}$ represent the key and value derived from the original audio features $F_a$. 

In parallel to AVE, we use the VAE to iteratively update the audio features with multi-scale features as guidance, the process of which is summarized as follows: 
\begin{equation}
F^{i-1}_{v\to a}=LN(Softmax(\frac{Q^{i-1}_{a}K^{i^{\top}}_{v}}{\sqrt{d_{head}}})V^{i}_{v}+F^{i-1}_{a}),
\end{equation}
\begin{equation}
F^{i}_{a}=LN(FFN(F^{i-1}_{v\to a})+F^{i-1}_{v\to a}).
\end{equation}
Here, ${F}^{i}_{a} \in \mathbb{R}^{T \times d}, i\in{\{1, 2, 3, 4\}}$ represents the audio features that have been enhanced by the visual features from the $i$-th scale. The term $F^{i-1}_{v\to a}$ denotes the intermediate variable that facilitates this enhancement. Additionally, $Q^{i-1}_{a}$ refers to the query generated from $F^{i-1}_{a}$, and $K^{i}_{v}$ and $V^{i}_{v}$ represent the key and value, respectively, derived from $F^{i}_{v}$.
The final enhanced audio features $F^{4}_{a}$ are then reshaped into $F^{'}_{a}\in{\mathbb{R}^{T\times{1}\times{d}}}$ and sent to the Attention Query Generator for interaction with intra-frame queries.
 
\subsection{Multi-query Transformer Module}

The EIM enables the dense interaction of audio-visual features. Building upon this foundation, we further utilize the Multi-query Transformer Module to capture the global embedding of sounding objects and promote frame-level and temporal-level interactions in a progressive process. 
The initial process involves employing a Transformer encoder for multi-scale feature fusion and refinement. Following this, we propose the Attention Query Generator, which leverages an attention mechanism to endow audio queries with dynamic learning capabilities, integrating these features into the intra-frame queries. These queries, in conjunction with visual features, are then input into the Multi-query Transformer Decoder to facilitate both intra- and inter-frame interactions. Specifically, intra-frame queries first interact with multi-scale visual features within the same frame. Then, inter-frame queries draw on spatial and temporal contextual information from the intra-frame interactions to enhance the perception of global features.
This staged processing culminates in the combination of the two types of queries, effectively capturing the object specificity within a single frame and generality across multiple frames.

\paragraph{Transformer Encoder}

Transformer Encoder is introduced to refine and integrate multi-scale visual features. To be specific, we gather and flatten the visual features from the last three scales of the updated features
$F_{v'}^i, i=\{2,3,4\}$ and then concatenate them as input queries $F_{\hat{v}}\in{\mathbb{R}^{\sum_{i=2}^{4}{H_{i}W_{i}\times{d}}}}$ (where $H_{i}$ and $W_{i}$ denote the height and width of visual features from $i$-th scale). 
To maintain computational efficiency, we feed the visual queries into the Deformable Transformer Encoder\cite{zhu2020deformable}, enabling our model to adaptively select regions of interest for self-attention calculation. The extracted multi-scale visual features are fed into the segmentation head and the decoder simultaneously.
Consistent with previous works on image segmentation~\cite{cheng2022masked,li2023mask}, 
we preserve the 1/8 feature maps $F_{v'}^1$ and directly take them as input to the segmentation head.

\paragraph{Attention Query Generator}
Different from previous approaches that obtain audio queries either through random initialization or by repeating audio features ~\cite{li2023catr,huang2023discovering,liu2023audio}, our approach employs an Attention Query Generator. This innovation enhances the intra-frame audio queries, facilitating dynamic learning of multimodal features. Specifically, we initiate the audio queries within each frame, denoted as $Q \in \mathbb{R}^{T \times n \times d}$, through random initialization, where \( n \) denotes the number of queries. 
Specifically, the audio queries decompose frame-level instance information from sparse audio features $F'_{a}$ using cross-attention layers. Meanwhile, the self-attention layers facilitate continuous interaction and updating of these features.
This process of the \( l \)-th layer can be formulated as follows:
\begin{equation}
Q^{l'}=LN(MHSA(Q^{l})+Q^{l})
\end{equation}
\begin{equation}
Q^{l+1}=LN(MHCA(Q^{l'}, F'_{a})+Q^{l'}),
\end{equation}
where $MHSA$ and $MHCA$ represent the multi-head self-attention and cross-attention, respectively. $Q^{l}$ refers to the queries of the $l$-th layer and $Q^{l'}$ represents the intermediate audio queries. 

\paragraph{Multi-query Transformer Decoder}
Prior researches focus more on intra-frame feature interactions, while the modeling of cross-modal interactions over temporal dimensions has not been adequately addressed. To bridge this gap, we introduce the Multi-query Transformer Decoder, which performs Intra-frame Cross-modal Interaction (ICI) and Inter-frame Temporal Interaction (ITI) on audio-visual features.

On the one hand, the intra-frame queries are used to capture the specialty of sounding objects within a single frame. 
As shown in Fig.~\ref{fig:2}, we employ the Deformable Transformer Decoder to facilitate intra-frame interactions between the intra-frame queries and the multi-scale feature maps, allowing the frame-wise object queries better access to the visual features of the sounding areas. 
The intra-frame queries $Q_{intra}\in{\mathbb{R}^{T\times{n}\times{d}}}$ are obtained by adding the output from the attention query generator and a set of learnable queries, each of them corresponding to an instance within the respective frames. In this module, the multi-scale visual features, acting as both key and value, dynamically update the queries through attention calculation with intra-frame spatial features, promoting the perception of the position and semantic information of sounding objects.

On the other hand, we exploit the inter-frame queries in ITI to capture the generality of sounding objects across temporal sequences. To better integrate local and global information for enhanced performance, we insert several layers of the ITI into each layer of the ICI as illustrated in Fig.~\ref{fig:2}. 
Specifically, the intra-frame queries are reshaped into $\hat{Q}_{intra}\in{\mathbb{R}^{nT\times{d}}}$ and fed into the self-attention layers to enhance the intra-frame queries' correspondence, which is denoted as: 
\begin{equation}
\hat{Q}^{'}_{intra}=LN(MHSA(\hat{Q}_{intra})+\hat{Q}_{intra}).
\end{equation}

\begin{figure}
    \centering
    \vspace{-2mm}
    \includegraphics[width=1\linewidth]{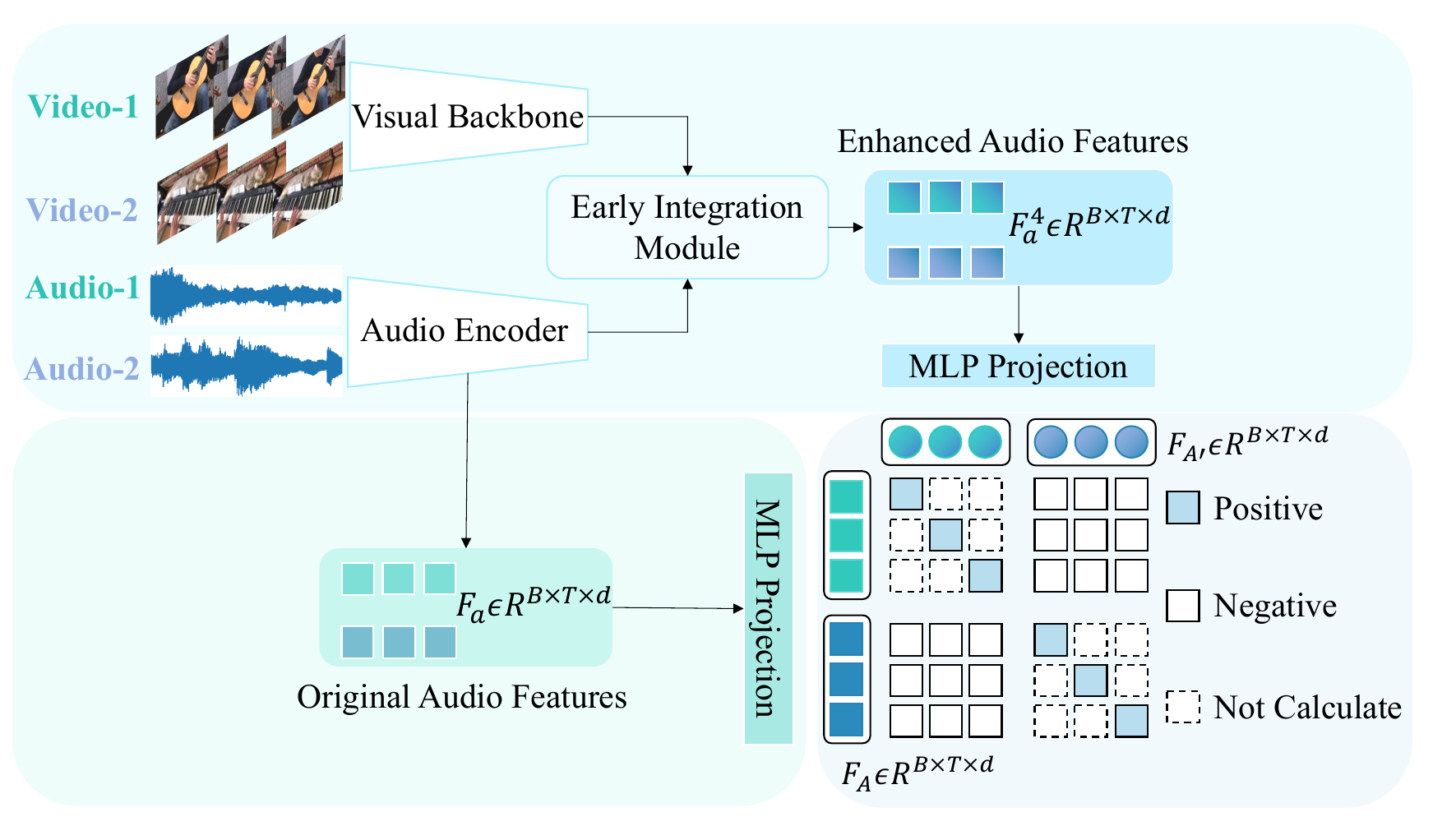}
    \vspace{-5mm}
    \caption{Bi-modal Contrastive Learning configuration. The contrastive loss is employed to compare the original audio features with the audio features fused with visual information. We consider the features of corresponding audio frames as positive (the blue blocks) and regard the audio frames from different audio samples within the same batch as negative (the solid white border).}
    \vspace{-1mm}
    \label{fig:4}
\end{figure}

Then, the updated intra-frame queries $\hat{Q}^{'}_{intra}$ interact with randomly initialized inter-frame queries $Q_{inter}\in{\mathbb{R}^{{n}\times{d}}}$ via a couple of cross-attention and self-attention layers, represented as:
\begin{equation}
    Q_{inter}^{l'}=LN( MHCA({Q}^{l}_{inter}, \hat{Q}^{'}_{intra})+Q_{inter}^{l}),
\end{equation}
\begin{equation}
Q^{l+1}_{inter}=LN(MHSA(Q^{l'}_{inter})+Q^{l'}_{inter}),
\end{equation}
where $Q^{l}_{inter}$ represents the inter-frame queries for the $l$-th ITI layer and $Q_{inter}^{l'}$ denotes the intermediate variable. 

By incorporating the ITI within ICI seamlessly and performing interaction interactively by $N\times M$ times, the visual features, audio features and hierarchical queries are fully incorporated.  
Finally, we repeat the output inter-frame queries $Q^{f}_{inter}\in{\mathbb{R}^{{n}\times{d}}}$ along the temporal dimension to obtain the object queries $Q^{f'}_{inter}\in{\mathbb{R}^{T\times{n}\times{d}}}$ for each frame, each of them representing the identical object across all the frames. To exploit the complementary of both types of queries, we combine them by the following equation:

\begin{equation}
    Q_{segment}=\hat{Q}^{f}_{intra}+Q^{f'}_{inter},
\end{equation}
where $\hat{Q}^{f}_{intra}\in{\mathbb{R}^{T\times{n}\times{d}}}$ represents the output intra-frame queries and $Q_{segment}$ represents the queries that will be used to generate the segmentation masks. 

\subsection{Segmentation Head}
The design of our segmentation head is inspired by ReferFormer~\cite{wu2022language}.
We integrate the original audio features $F_{a}$, the 1/8 feature maps $F^{1}_{v^{'}}$, and the Transformer encoder output into Cross-modal feature pyramid network (FPN). Employing spatial-reduction attention, we progressively integrate audio features into each scale of the visual features. Through this integration and subsequent upsampling, the mask feature $\mathcal{F} \in \mathbb{R}^{T \times \frac{H}{8} \times \frac{W}{8} \times d}$ is generated.
Then, the intermediate segmentation masks $\mathcal{M}_{s}\in{\mathbb{R}^{T\times{\frac{H}{8}}\times{\frac{W}{8}}\times{n}}}$ are generated by applying dynamic convolution kernels, predicted by \(Q_{segment}\), to the mask features.
Finally, we utilize a multi-layer perception to project the intermediate segmentation masks to \(\mathcal{M}^{'}_{s}\in{\mathbb{R}^{T\times{\frac{H}{8}}\times{\frac{W}{8}}\times{d}}}\). A fully connected layer is applied to predict a \(n_{class}\) segmentation map $\mathcal{M}$. The detailed processes involved are as follows:
\begin{equation}
\mathcal{M}=FC(MLP(\mathcal{M}_{s})+\mathcal{F}),
\end{equation}
where \(FC\) denotes the fully connected layer, \(\mathcal{M}\in{\mathbb{R}^{T\times{\frac{H}{8}}\times{\frac{W}{8}}\times{n_{class}}}}\) represents the final mask and $MLP$ represents the multi-layer perception.
\subsection{Bi-modal Contrastive Learning}
To align visual and audio features within a unified representational space, we introduce a Bi-modal Contrastive Learning (BCL) strategy, which leverages contrastive loss to enhance the correlation between these multimodal features. 
In contrast to prior approaches that utilize video-level contrastive losses~\cite{mo2023weakly,mo2023audio}, where global audio and visual features from the same sample are treated as positive and those from different samples as negative, our method introduces a frame-level contrastive loss. By doing so, our method takes into account the semantic information of the sounding objects across different audio frames.
Given that the sounding objects in different frames of the same audio clip can be identical, we designate corresponding audio frames as positive while treating audio frames from different samples within the same batch as negative.

As depicted in Fig.~\ref{fig:4}, BCL is performed between visually enhanced audio features $F^{4}_{a}$  and the original audio features $F_{a}$. 
Specifically,  for each batch containing $T$-length sequences of audio samples, we project both $F_{a}$  and $F^{4}_{a}$ into a common representation space. This results in two sets: $F_{A}=\left\{F_{A}^{bt}\right\}_{b=1,t=1}^{B,T}$ and $F_{A'}=\left\{F_{A'}^{bt}\right\}_{b=1,t=1}^{B,T}$, where $F_{A}^{bt}$ and $F_{A'}^{bt}$ are vectors in $\mathbb{R}^{d}$, and $B$ denotes the batch size. The formulation of our contrastive loss $\mathcal{L}_{bi}$ is represented as follows: 
\begin{equation}
\begin{aligned}
\mathcal{L}_{bi}=-\frac{1}{BT}&\sum_{b=1}^{B}\sum_{t=1}^{T}
(log{\frac{exp(\frac{1}{\tau}S(F_{A}^{bt},F_{A'}^{bt}))}{\sum_{i\neq{b}}\sum_{j=1}^{T}exp(\frac{1}{\tau}S(F_{A}^{bt},F_{A'}^{ij}))}} \\
+&log{\frac{exp(\frac{1}{\tau}S(F_{A'}^{bt},F_{A}^{bt}))}{\sum_{i\neq{b}}\sum_{j=1}^{T}exp(\frac{1}{\tau}S(F_{A'}^{bt},F_{A}^{ij})}}
),
\end{aligned}
\end{equation}
where $\tau$ represents the temperature constant and $S$ signifies the similarity of features.
Through the minimization of contrastive loss, different modalities achieve alignment and convergence within a unified feature space.

In addition to the bi-modal contrastive loss, our approach employs the dice loss~\cite{milletari2016v} for supervising the output masks at each layer. The cumulative loss function is formulated as follows:
\begin{equation}
\mathcal{L}_{total}=\mathcal{L}_{dice}+\lambda\mathcal{L}_{bi},
\end{equation}
where $\lambda$ represents the loss trade-off. 

\section{Experiments}
\subsection{Datasets and Metrics}
\paragraph{Datasets} 
The training and evaluation are performed on two public benchmarks: AVSBench-object~\cite{zhou2022audio} and AVSBench-semantic~\cite{zhou2023audio}.
AVSBench-object consists of 5 synchronized audio-video pairs, sampled at a frequency of 1 frame per second. It comprises two settings: the semi-supervised Single Sound Source Segmentation (S4) and the fully supervised Multiple Sound Source Segmentation (MS3). 
S4 consists of 4932 videos across 23 categories, each featuring single consistent sound source throughout the entire continuous scene, with annotations provided only for the first frame. 
MS3 comprises 424 videos, incorporating complex transitions of multiple-sounding objects, with each frame annotated by the binary masks. 
AVSBench-semantic, on the other hand, offers a broader range of 12356 videos spanning 70 categories. Each video is
either 5 or 10 frames in length and annotated with the semantic-level mask to classify the objects depicted. 
\paragraph{Metrics}
We use F-score ($M_{\mathcal{F}}$) and Jaccard index ($M_{\mathcal{J}}$) to evaluate the performance of our model. $M_{\mathcal{J}}$ quantifies the intersection-over-union between the prediction and the ground-truth mask. $M_{\mathcal{F}}$ is defined as $\frac{(1+\beta^{2})\times{precision}\times{recall}}{\beta^{2}\times{precision}+recall}$ (where $\beta$ is set as 0.3 in line with TPAVI~\cite{zhou2022audio}), measuring the balance between precision and recall. 

\definecolor{lightskyblue}{rgb}{0.53, 0.81, 0.98}
\begin{table}[!tbp]
  \centering
  \caption{Comparison with state-of-the-art methods on the MS3 and S4 settings.}
  \label{tab:comparison with others in AVSBench-object}
  \vspace{-2mm}
  \scalebox{0.99}
  {
  \begin{tabular}{p{2.2cm}<{\raggedright} p{1.4cm}<{\raggedright} p{0.6cm}<{\centering} p{0.6cm}<{\centering} p{0.6cm}<{\centering} p{0.6cm}<{\centering}}
    \toprule[1.2pt]
    \multirow{2}{*}{Method}& \multirow{2}{*}{Backbone} & \multicolumn{2}{c}{S4}& \multicolumn{2}{c}{MS3}\\
     &  & $M_\mathcal{J}$ & $M_\mathcal{F}$ & $M_\mathcal{J}$ & $M_\mathcal{F}$ \\
    \midrule
    \multirow{2}{*}{TPAVI (ECCV'22)~\cite{zhou2022audio}} &  ResNet-50 & 72.8 & 84.8   & 47.9 & 57.8   \\
     &  PVT-v2 & 78.7 & 87.9 & 54.0 & 64.5 \\
    \midrule
    \multirow{2}{*}{CATR (ACMMM'23)~\cite{li2023catr}} & ResNet-50  & 74.8 & 86.6 & 52.8 & 65.3 \\
     & PVT-v2 & 81.4 & 89.6 & 59.0 & 70.0 \\
    \midrule
    \multirow{2}{*}{AQFormer ($\textup{IJCAI}$'23)~\cite{huang2023discovering}} & ResNet-50  & 77.0 & 86.4 & \underline{55.7} & \underline{66.9} \\
     & PVT-v2 & 81.6 & 89.4 & 61.1 & 72.1 \\
    \midrule
    \multirow{2}{*}{AVSBiGen ($\textup{AAAI}$'24)~\cite{hao2023improving}} & ResNet-50 & 74.1 & 85.4 & 45.0 & 56.8\\
     & PVT-v2 & 81.7 & 90.4 & 55.1 & 66.8 \\
    \midrule
     \multirow{2}{*}{ECMVAE ($\textup{ICCV}$'23)~\cite{mao2023multimodal}} & ResNet-50 & 76.3 & 86.5 & 48.7 & 60.7\\
     & PVT-v2 & 81.7 & 90.1 & 57.8 & 70.8\\
    \midrule
    \multirow{2}{*}{DIFFAVS ($\textup{Arxiv}$'23)~\cite{mao2023contrastive}}&ResNet-50& 75.8 & 86.9 & 49.8 & 58.2\\
    & PVT-v2 & 81.4 & 90.2 & 58.2 & 70.9\\
    \midrule
    \multirow{2}{*}{AVSegFormer ($\textup{AAAI}$'24)~\cite{gao2024avsegformer}} & ResNet-50 & 76.4 & 86.7 & 53.8 & 65.6 \\
     & PVT-v2 & 83.1 & 90.5 & 61.3 & 73.0 \\
    \midrule
    \multirow{2}{*}{BAVS ($\textup{TMM}$'24)~\cite{liu2024bavs}} &ResNet-50 & \underline{78.0} & 85.3 & 50.2 & 62.4\\
    &PVT-v2& 82.0 & 88.6 & 58.6 & 65.5\\
    \midrule
     \multirow{2}{*}{AVSAC ($\textup{Arxiv}$'24)~\cite{chen2024bootstrapping}} & ResNet-50 & 76.9 & \underline{87.0} & 54.0 & 65.8\\
     & PVT-v2 & \underline{84.5} & \underline{91.6} & \underline{64.2} & \underline{76.6}\\
    \midrule
    \rowcolor{gray!20}
      &  ResNet-50  & \textbf{79.4}& \textbf{90.0}& \textbf{64.7}& \textbf{74.3}\\
      \rowcolor{gray!20}
    \multirow{-2}{*}{CCFormer}& PVT-v2 & \textbf{84.9} & \textbf{92.7} & \textbf{70.7}& \textbf{80.5}\\
    \bottomrule[1.2pt]
  \end{tabular}
}
\vspace{-2mm}
\end{table}

\subsection{Implementation Details}
We utilize ResNet-50~\cite{he2016deep} pretrained on the MSCOCO dataset~\cite{lin2014microsoft} and PVT-v2~\cite{wang2022pvt} pretrained on the ImageNet dataset~\cite{russakovsky2015imagenet} as the visual backbones. This configuration aligns with the setups employed in prior studies~\cite{zhou2022audio,huang2023discovering}. For the audio encoder, we utilize VGGish~\cite{hershey2017cnn} pretrained on the AudioSet~\cite{gemmeke2017audio}.  Both intra-frame and inter-frame queries are configured to five, and three layers are employed in each of the Transformer Encoder and Decoder to achieve optimal performance.
The training batch sizes are tailored to each dataset, with two for AVSBench-object and six for AVSBench-semantic. Hypermeters $\lambda$ is set to 0.1. We choose the AdamW optimizer~\cite{loshchilov2017decoupled}, initializing the learning rate at 2e-5. The training process spans 120 epochs for the MS3 setting, 60 epochs for the S4 setting, and 40 epochs for the AVSBench-semantic dataset. For AVSBench-object, we use the combination of dice loss and contrastive loss, while for AVSBench-semantic, we solely employ the per-pixel binary cross-entropy loss as the supervision signal.
Since the number of frames varies in the AVSBench-semantic dataset, we preprocess by padding all videos to a uniform length of 10 frames to facilitate easier handling.
In alignment with methodologies from previous studies~\cite{ying2023ctvis,yan2023referred,luo2023soc}, we apply a consistent data augmentation strategy across all datasets. This strategy includes horizontal flipping, photometric distortion, random resizing, and cropping. All training procedures are conducted on a single NVIDIA A800 GPU.

\begin{table}[tbp]
 \centering
  \caption{Comparison with state-of-the-art methods on the AVSBench-semantic dataset.}
  \label{tab:comparison with others in AVSS}
  \scalebox{1.05}
  {
  \begin{tabular}{p{1.8cm}<{\raggedright} p{1.3cm}<{\centering} p{0.5cm}<{\centering} p{0.5cm}<{\centering}}
    \toprule[1.2pt]
    Method & Backbone & $M_\mathcal{J}$ & $M_\mathcal{F}$ \\
    \midrule
    \multirow{2}{*}{TPAVI ($\textup{Arxiv}$'23)~\cite{zhou2023audio}}  &  ResNet-50 & 20.2 & 25.2  \\
     &  PVT-v2 & 29.8 & 35.2 \\
    \midrule
    \multirow{2}{*}{BAVS ($\textup{TMM}$'24)~\cite{liu2023bavs}}  & ResNet-50  & 24.7& 29.6\\
     & PVT-v2 & 32.6 & 36.4 \\
    \midrule
    \multirow{2}{*}{AVSAC ($\textup{Arxiv}$'24)~\cite{chen2024bootstrapping}} & ResNet-50 & 25.4& 29.7\\
     & PVT-v2 & 37.0& 42.4\\ 
    \midrule
    \multirow{2}{*}{AVSegFormer ($\textup{AAAI}$'24)~\cite{gao2024avsegformer}} & ResNet-50 & \underline{26.6}& \underline{31.5} \\
    & PVT-v2 & \underline{37.3} & \underline{42.8} \\
    \midrule
    \rowcolor{gray!20}
    & ResNet-50  & \textbf{32.0}& \textbf{38.1}\\
    \rowcolor{gray!20}
     \multirow{-2}{*}{CCFormer}& PVT-v2 & \textbf{42.3}& \textbf{48.5}\\
    \bottomrule[1.2pt]
    \end{tabular}
  }
\end{table}

\subsection{Quantitative Results} 
Table~\ref{tab:comparison with others in AVSBench-object} presents the performance comparison of our CCFormer against other leading methods~\cite{zhou2022audio,huang2023discovering,li2023catr,hao2023improving,gao2024avsegformer,liu2024bavs,chen2024bootstrapping,mao2023contrastive,mao2023multimodal} on the AVSBench-object dataset. For the S4 setting, our model demonstrates a significant advancement over the state-of-the-art methods, achieving gains of 1.4\% in \(M_{\mathcal{J}}\) and 3.0\% in \(M_{\mathcal{F}}\) with the ResNet-50 backbone and by 0.4\% \(M_{\mathcal{J}}\) and 1.1\% \(M_{\mathcal{F}}\) with the PVT-v2 backbone. This indicates that CCFormer enhances the perception of a single sound source persisting through the same scene by continuously modeling dynamic temporal information across multiple frames. 
In the MS3 setting, our model exhibits improvements of 9.0\% in \(M_{\mathcal{J}}\) and 7.4\% in \(M_{\mathcal{F}}\) using ResNet-50, and increases of 6.5\% in \(M_{\mathcal{J}}\) and 3.9\% in \(M_{\mathcal{F}}\) with PVT-v2. 
The improvement originates from the complementary framework combining intra- and inter-frame interactions, allowing the model to integrate local and global features to harness the scenarios involving multiple sound source transformations.

Table~\ref{tab:comparison with others in AVSS} illustrates the results of our model on the AVSBench-semantic dataset. In this more challenging semantic segmentation task, our method demonstrates a substantial lead, yielding improvements of 5.4\% in \(M_{\mathcal{J}}\) and 6.6\% in \(M_{\mathcal{F}}\) with ResNet-50, and 5.0\% in \(M_{\mathcal{J}}\) and 5.7\% in \(M_{\mathcal{F}}\) with PVT-v2. 
The results demonstrate the effectiveness of CCFormer in extracting and distinguishing semantic contents.  

\begin{table}[!tbp]
    \centering
    \caption{Comparison with previous methods on parameters, computational complexity and GPU memory with single video and audio frame at the resolution of $224\times{224}$. 
    }
    \vspace{-1mm}
    \label{tab: parameter analysis}
    \scalebox{0.9}{
    \begin{tabular}{ccccc}
    \toprule
    Methods & Params (M) & FLOPs (G) & FPS & GPU (MB) \\
    \midrule
    TPAVI (ECCV'22)~\cite{zhou2022audio} & 101.3 & 34.9 & 13.7 & 1324\\
    ECMVAE ($\textup{ICCV}$'23)~\cite{mao2023multimodal} & 91.3 & 19.9 & 11.2 & 1274\\
    AVSegFormer ($\textup{AAAI}$'24)~\cite{gao2024avsegformer} & 113.9 & 49.0 & 9.8 & 1364\\
    CATR (ACMMM'23)~\cite{li2023catr} & 115.8 & 22.3 & 10.6 & 1322 \\
    CCFormer & 143.4 & 86.3 & 7.8 & 1482\\
    \bottomrule
    \vspace{-3mm}
    \end{tabular}
    }
\end{table}
We analyse the total number of trainable parameters, computational efficiency and GPU memory of CCFormer against other methods in Table~\ref{tab: parameter analysis}.
It can be observed that CCFormer exhibits higher parameter count and computational complexity than previous open-sourced methods. Moreover, Table~\ref{tab: ablation study of all the modules} presents the number of parameters and FLOPs when core components are removed. It is evident that introducing inter-frame interactions increases the parameter count but also leads to significant performance improvements. 
Additionally, as demonstrated in Table~\ref{tab: ablation study of number of ICI layers} and ~\ref{tab:ablation study of the number of ITI layers}, further increasing the model’s parameter count actually leads to a decline in performance. This indicates that the improvement in model performance is not directly dependent on the increase in parameter count. We will provide a detailed explanation of the role of each critical component in Section~\ref{sec:ablation study}.

\begin{figure*}
    \centering
    \includegraphics[width=1\linewidth]{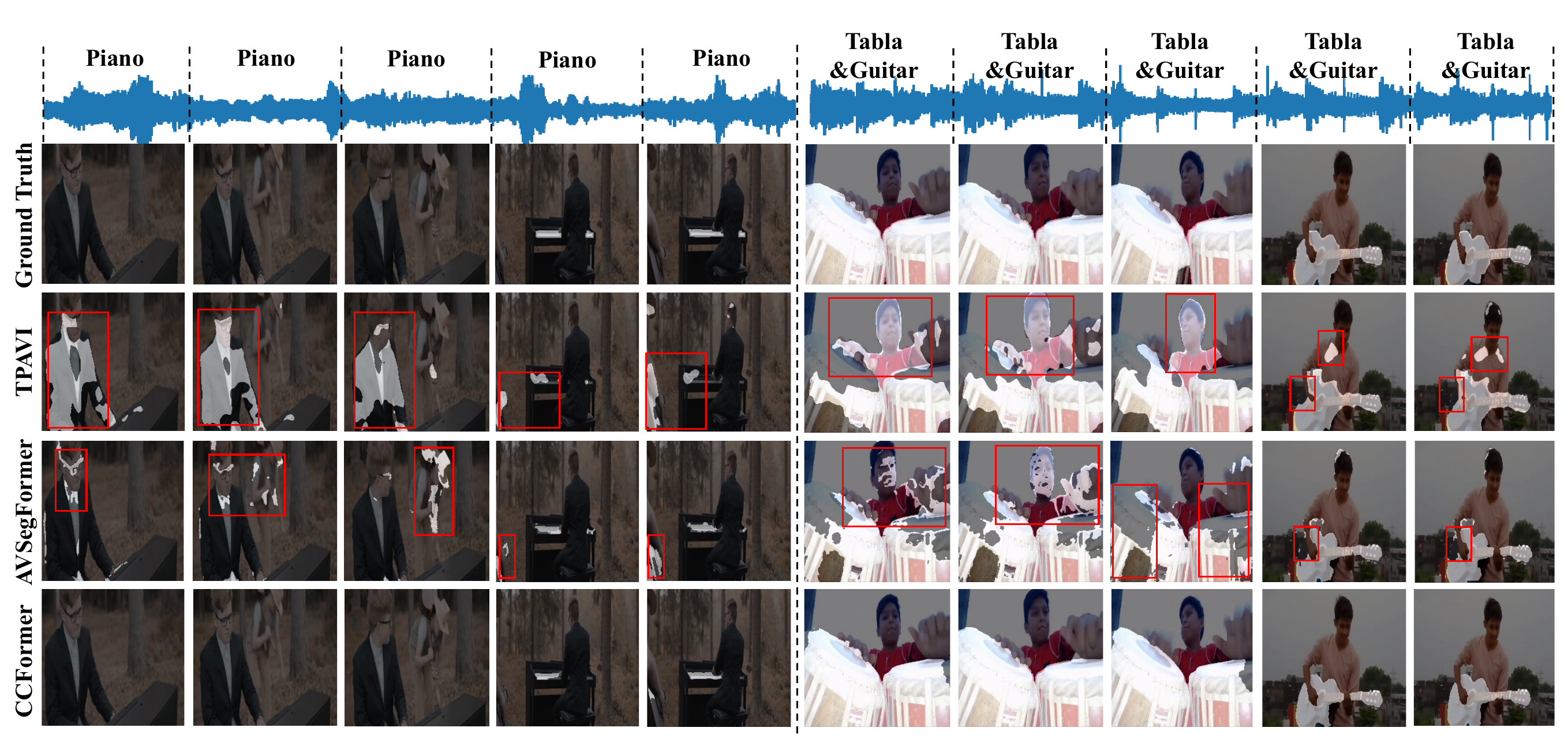}
\vspace{-5mm}
    \caption{Qualitative comparisons of TPAVI~\cite{zhou2022audio}, AVSegFormer~\cite{gao2024avsegformer} and our CCFormer on AVSBench-object dataset. We provide scenarios with scene transition.}
    \label{fig:ms3 comparison}
\end{figure*}
\begin{figure}
    \centering
    \vspace{-2mm}
    \includegraphics[width=1\linewidth]{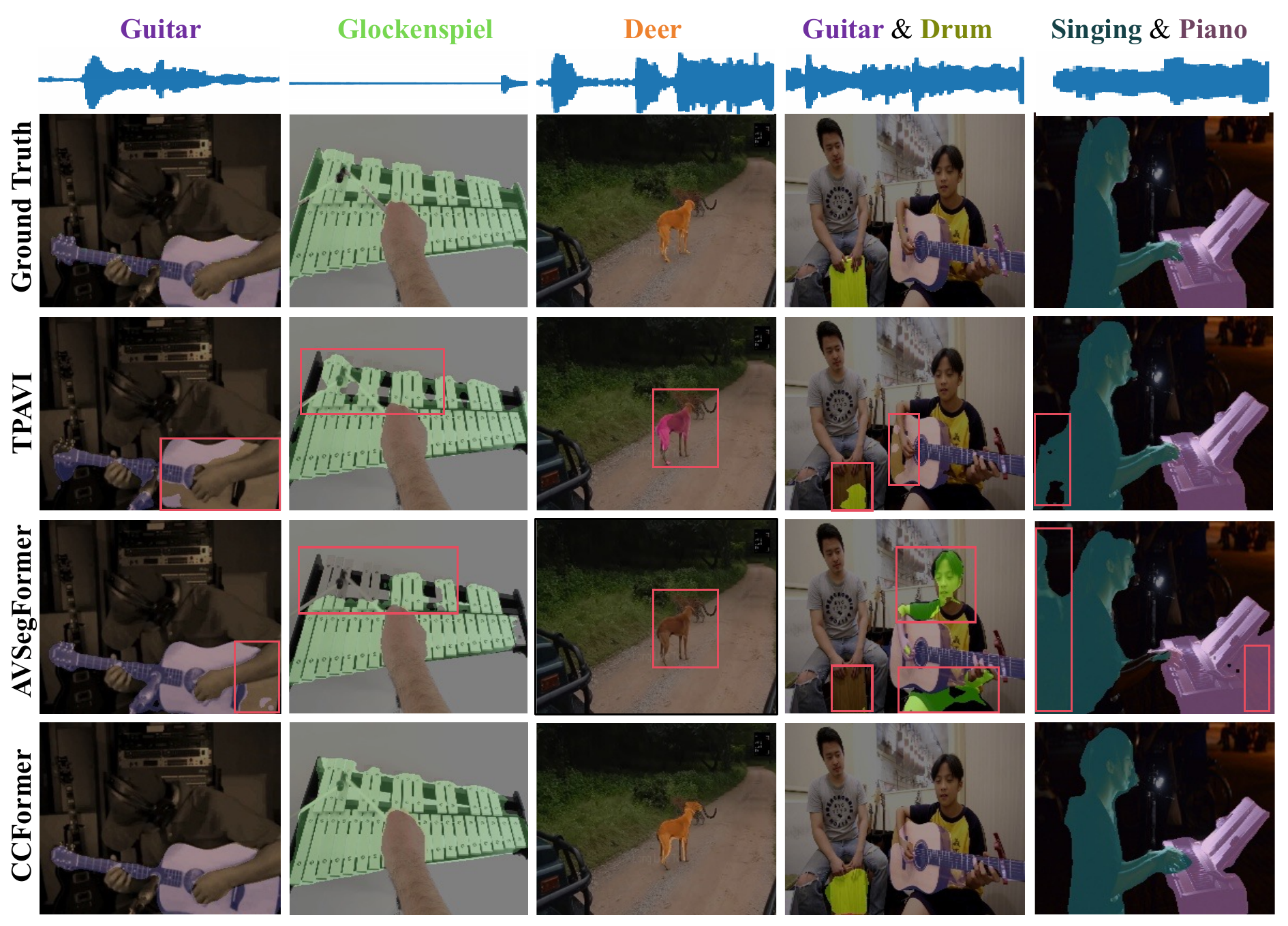}
\vspace{-5mm}
    \caption{Qualitative comparisons of TPAVI~\cite{zhou2022audio}, AVSegFormer~\cite{gao2024avsegformer} and our CCFormer
    on AVS-Semantic dataset. 
    }
\vspace{-1mm}
    \label{fig:avss comparison}
\end{figure}

\begin{figure}
    \centering
    \vspace{-2mm}
    \includegraphics[width=1\linewidth]{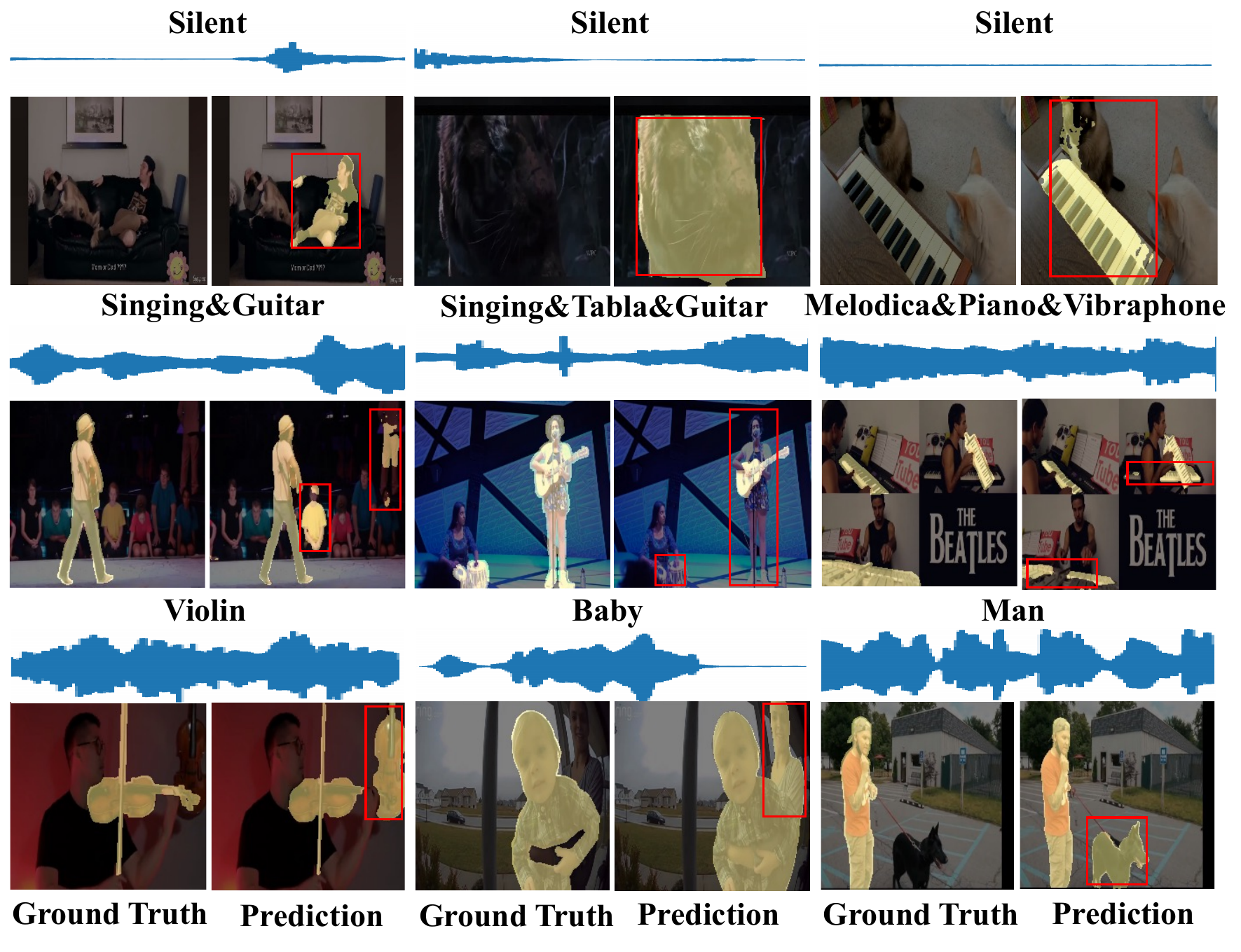}
    \vspace{-5mm}
    \caption{Failure cases analysis. We select several failure scenarios from MS3 setting, including silent frames, multiple sounding objects, and dynamic changes in sound sources. }
    \label{fig:failure cases}
\end{figure}

\subsection{Qualitative Results}

\paragraph{Comparison of AVSBench-object}
Fig.~\ref{fig:ms3 comparison} showcases the generated masks from various models utilizing the PVT-v2 backbone. The results demonstrate that previous methods fail to adequately adapt to situations where auditory objects are absent in certain video frames. For example, both TPAVI and AVSegFormer are inclined to segment the individual playing the piano when only piano sounds are given without visual presence. However, once the piano becomes visually apparent in the final two frames, neither can accurately identify it, mistakenly isolating the guitar located at the edge instead.
Similarly, in the right video sequence, these methods initially fail to delineate the boundaries of the tabla accurately, incorrectly segmenting the boy in the first three frames. Additionally, the masks generated in the final two frames exhibit incompleteness and discontinuity. In contrast, CCFormer effectively utilizes inter-frame queries to model temporal consistency, which enables it to extract information across continuous frames and achieve precise segmentation of sounding objects.

\paragraph{Comparison of AVSBench-semantic}
Fig.~\ref{fig:avss comparison} illustrates audio-visual semantic segmentation (AVSS) results on AVSBench-semantic, with different colors representing various categories. CCFormer significantly outperforms prior methods, demonstrating superior ability in extracting semantic information without the need for specific designs tailored for semantic segmentation. In the first two columns, previous methods, TPAVI and AVSegFormer, produce segmentation outcomes characterized by discontinuities and holes. In contrast, CCFormer generates more refined masks for auditory instances. In the third case, TPAVI incorrectly classifies a deer as the sound source, whereas AVSegFormer processes it as a silent frame. Moving to the fourth scenario, TPAVI fails to completely segment the sounding drum, and AVSegFormer incorrectly attributes the sound source to the guitar player. Similarly, in the final column, the segmentation masks of both methods incorporate redundant parts beyond the sound sources. In all these cases, CCFormer consistently excels, accurately identifying and classifying sound sources with impressive semantic accuracy.

\begin{table}[tbp]
\centering
\caption{Comprehensive analysis of overall components.
}
\vspace{-1mm}
    \label{tab: ablation study of all the modules}
    \scalebox{0.9}{
        \begin{tabular}{ccccccc}
        \toprule
        \multirow{2}{*}{Components}& \multicolumn{2}{c}{MS3}& \multicolumn{2}{c}{S4} & \multirow{2}{*}{Params (M)}& \multirow{2}{*}{FLOPs (G)}\\
             &\(M_{\mathcal{J}}\)&\(M_{\mathcal{F}}\)& \(M_{\mathcal{J}}\)&\(M_{\mathcal{F}}\)\\
             \midrule
            w/o EIM & 67.7& \underline{78.5} & \underline{83.8} & \underline{92.3}& 
            140.8 & 80.8\\
            w/o ICI Module& 68.1& 77.5 & 83.2& 91.9& 138.8 & 85.9\\
            w/o ITI Module & 67.4 & 77.8 & 83.1 & 91.8 & 117.3 & 86.1\\
            w/o BCL & \underline{69.3}& 78.3 &83.7 & 92.1& 143.0 & 85.9\\
            CCFormer& \textbf{70.7}&\textbf{80.5} & \textbf{84.9}& \textbf{92.7}& 143.4 & 86.3\\
        \bottomrule
        \end{tabular}
        }
        \vspace{-3mm}
\end{table}
\begin{table}[!tbp]
    \centering
    \caption{Ablation study of the design of query generator. 
    }
    \vspace{-1mm}
    \label{tab: ablation study of query generator}
    \begin{tabular}{ccccc}
    \toprule
         \multirow{2}{*}{Query Generator}& \multicolumn{2}{c}{MS3}& \multicolumn{2}{c}{S4}\\
         &\(M_{\mathcal{J}}\)&\(M_{\mathcal{F}}\)&\(M_{\mathcal{J}}\)&\(M_{\mathcal{F}}\)\\
         \midrule
         w/o Query Generator&  68.1& 76.6& 84.0&\underline{92.4}\\
         Repeat Generator& \underline{68.9}&\underline{78.1}& \underline{84.2}& 92.3\\
         Attention Generator& \textbf{70.7}&\textbf{80.5}&\textbf{84.9}&\textbf{92.7}\\
    \bottomrule
    \vspace{-3mm}
    \end{tabular}
\end{table}
\begin{table}[!tbp]
\centering
    \caption{Ablation results of the Early integration module. 
    }
    \label{tab:ablation study of early integration module}
        \begin{tabular}{cccccc}
        \toprule
        \multirow{2}{*}{AVE}&\multirow{2}{*}{VAE}&\multicolumn{2}{c}{MS3}&\multicolumn{2}{c}{S4}\\
        & & \(M_{\mathcal{J}}\)&\(M_{\mathcal{F}}\)&\(M_{\mathcal{J}}\)&\(M_{\mathcal{F}}\)\\
             \midrule
             & &  67.7& 78.5 & 83.8 & 92.3\\
             &{$\surd$}&  68.1& 78.3&84.2  &92.3 \\
             {$\surd$}&&  \underline{69.1}&  \underline{79.6}& \underline{84.5} & \underline{92.4}\\
             {$\surd$}&{$\surd$}& \textbf{70.7}&\textbf{80.5}& \textbf{84.9}& \textbf{92.7}\\
        \bottomrule
        \end{tabular}   
\end{table}

\subsection{Failure Cases Analysis}
We conduct a detailed analysis of several failure cases, as illustrated in Fig.~\ref{fig:failure cases}. These cases are selectively drawn from the MS3 dataset and represent a diverse array of scenarios, including silent frames, scenes with multiple sounding objects, and transitions involving dynamic sound sources.

In our analysis, we identify specific challenges encountered by CCFormer in processing audio-visual data.
In the first row, CCFormer erroneously identifies silent frames as containing sounding objects. In the second row, when video frames contain numerous obscured sound sources, CCFormer fails to accurately capture the correct sound-producing objects, resulting in missed targets and incomplete masks. The third row involves dynamically changing sound sources while the scenes remain similar in successive frames. Here, CCFormer may introduce target features from adjacent frames during inter-frame interactions, leading to a failure in properly identifying sound-producing objects in the current frame. These instances underscore the potential for further refinement of CCFormer, which combines intra- and inter-frame integration. 

\subsection{Ablation Study}  
\label{sec:ablation study}
We conduct ablative experiments under both the MS3 and S4 settings, utilizing PVT-v2 as the visual backbone in all cases. Comprehensive experimental results are provided in the following sections.

\begin{table}[!tbp]
\centering
    \caption{Ablation study of the number of ICI layers.
    }
    \label{tab: ablation study of number of ICI layers}
    \begin{tabular}{ccccc}
    \toprule
    \multirow{2}{*}{Num. of \ Layers}&\multicolumn{2}{c}{MS3}&\multicolumn{2}{c}{S4}\\
    &\(M_{\mathcal{J}}\)&\(M_{\mathcal{F}}\)&\(M_{\mathcal{J}}\)&\(M_{\mathcal{F}}\)\\
        \midrule
         1& 68.5&78.4& 83.8& 92.4\\
         2&  \underline{70.3}& \underline{79.6}& \underline{84.2}& \underline{92.5}\\
         3& \textbf{70.7}& \textbf{80.5}& \textbf{84.9}& \textbf{92.7}\\
         4& 69.6& 79.5 & 83.8& 92.1\\
    \bottomrule
    \end{tabular}
    \vspace{-3mm}
\end{table}
\begin{table}[tbp]
\centering
    \caption{Ablation analysis of the number of ITI layers.
    }
    \label{tab:ablation study of the number of ITI layers}
    \begin{tabular}{ccccc}
    \toprule
    \multirow{2}{*}{Num. of \ Layers}&\multicolumn{2}{c}{MS3}&\multicolumn{2}{c}{S4}\\
    &\(M_{\mathcal{J}}\)&\(M_{\mathcal{F}}\)&\(M_{\mathcal{J}}\)&\(M_{\mathcal{F}}\)\\
         \midrule
         1&  67.8& 78.2& 83.7 & 92.1\\
         2&\underline{69.2}&79.7& \underline{84.2}& 92.4\\
         3&\textbf{70.7}&\textbf{80.5}& \textbf{84.9}& \textbf{92.7}\\
         4&68.5&\underline{79.8}& 83.9 & \underline{92.5}\\
 \bottomrule
 \end{tabular}
 \vspace{-3mm}
\end{table}
\begin{table}[!tbp]
\centering
    \caption{Ablation Study of the Bi-modal Contrastive Learning. 
    }
    \label{tab:ablation study of bcl}
    \begin{tabular}{ccccc}
    \toprule
    \multirow{2}{*}{Another modality}& \multicolumn{2}{c}{MS3} & \multicolumn{2}{c}{S4}\\
         &\(M_{\mathcal{J}}\)&\(M_{\mathcal{F}}\)&\(M_{\mathcal{J}}\)&\(M_{\mathcal{F}}\)\\
         \midrule
         Ori. visual feat.&  \underline{69.5}& 79.4& 83.9 & 92.4\\
         Enh. visual feat.&  69.2& \underline{79.5}& \underline{84.3}& \underline{92.6}\\
         Enh. audio feat. & \textbf{70.7}& \textbf{80.5}& \textbf{84.9}& \textbf{92.7}\\
    \bottomrule
    \end{tabular}
    \vspace{-3mm}
\end{table}

\paragraph{Overall Component Analysis}
We examine the influence of specific components on the model's performance. 
This includes individual analyses of the EIM, ICI, ITI and BCL, with results presented in Table~\ref{tab: ablation study of all the modules}. 
These results indicate that the inclusion of inter-frame interactions yields improvements of 3.3\% $M_{\mathcal{J}}$ in MS3 and 1.8\% $M_{\mathcal{J}}$ in S4, highlighting the importance of integrating temporal-level information. 
It can also be observed that cross-modal interaction within single frame facilitates the utilization of spatial features, leading to substantial improvements of 2.6\% $M_{\mathcal{J}}$ in MS3 and 1.7\% $M_{\mathcal{J}}$ in S4. 
 In addition, incorporating early fusion strategy brings performance gains of 3.0\% $M_{\mathcal{J}}$ in MS3 and 1.1\% $M_{\mathcal{J}}$ in S4, underscoring the necessity of balanced modal information at the initial stage.  
Furthermore, the integration of BCL contributes to a notable increase of 1.4\% $M_{\mathcal{J}}$ in MS3 and 1.2\% $M_{\mathcal{J}}$ in S4. These findings robustly support the effectiveness of the proposed modules.

\paragraph{Query Generator Analysis}
The effectiveness of the Attention Query Generator is analyzed, with experimental results presented in Table~\ref{tab: ablation study of query generator}.
We observe that, compared to random initialization or directly replicating audio features, utilizing the attention mechanism to generate intra-frame queries can not only integrate audio features, but also enhance the perceptual capability of audio queries for auditory signals, culminating in superior segmentation outcomes. 
\begin{figure}
    \centering
    \vspace{-2mm}
    \includegraphics[width=1\linewidth]{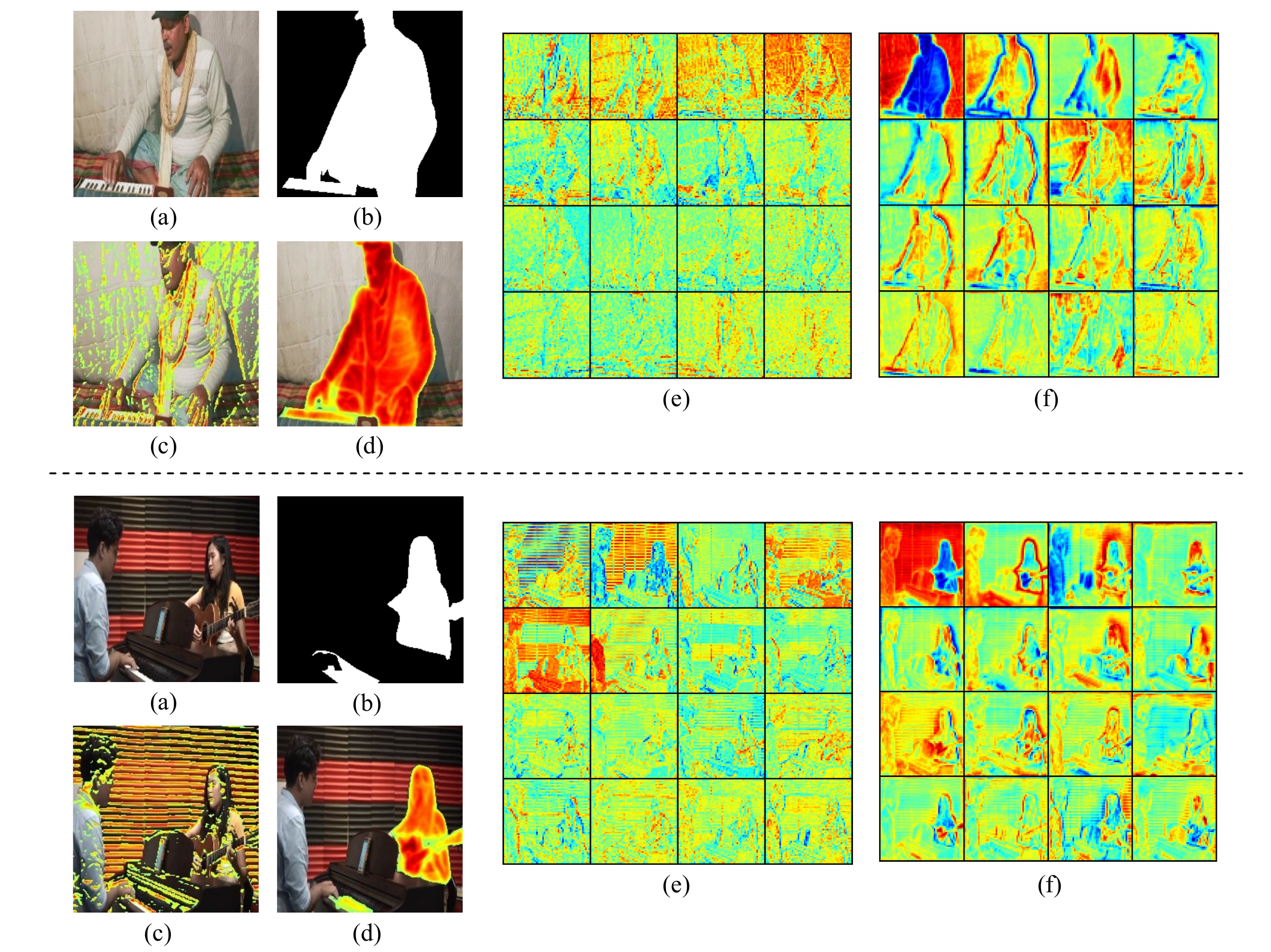}
    \vspace{-5mm}    \caption{Visualization of feature maps from visual backbones and after Cross-modal FPN. 
    (a) Raw images. (b) Ground truths. (c) Feature heatmaps from visual backbones. (d) Feature heatmaps after Cross-modal FPN. (e) Channel visualizations of feature maps from visual backbones and (f) after Cross-modal FPN.}
    \vspace{-1mm}
    \label{fig:feature map visualizaiton}
\end{figure}
\paragraph{Early Integration Module Analysis}
Table~\ref{tab:ablation study of early integration module} illustrates the influence of the Early Integration Module. The experimental results demonstrate the effectiveness of the two sub-modules. 
This is because injecting temporal-spatial information from visual features into audio features facilitates the audio queries to pay more attention to the auditory objects. 
Similarly, merging audio features into visual features assists in effectively filtering out the background information. 
\paragraph{Number of ICI and ITI layers} 
Further, we explore the optimal number of layers in the Multi-query Transformer. Table~\ref{tab: ablation study of number of ICI layers} presents the effects of varying the number of  ICI layers, while Table~\ref{tab:ablation study of the number of ITI layers} focuses on the impact of the number of ITI layers. 
The experimental results demonstrate that a configuration of three ICI layers, each followed by three ITI layers, achieves the best performance.  This finding suggests that an appropriate increase in Transformer layers enhances the extraction and interaction of both local and global information.  Nonetheless, an overabundance of ICI and ITI modules can increase the model's complexity excessively, potentially resulting in diminished performance.
\paragraph{Bi-modal Contrastive Learning Analysis}
Finally, we conduct an ablation analysis on BCL, as presented in Table~\ref{tab:ablation study of bcl}. Experiments are performed in which the enhanced audio features are replaced with spatially averaged pooled visual features from the largest scale. Substituting enhanced audio features with either original or enhanced visual features led to a decline in performance. This observation suggests that directly applying contrastive learning to both visual and audio modalities fails to effectively utilize the rich spatial information of the visual features, leading to insufficient alignment between the representations of the modalities.

\begin{figure}
    \centering
    \vspace{-2mm}
    \includegraphics[width=1\linewidth]{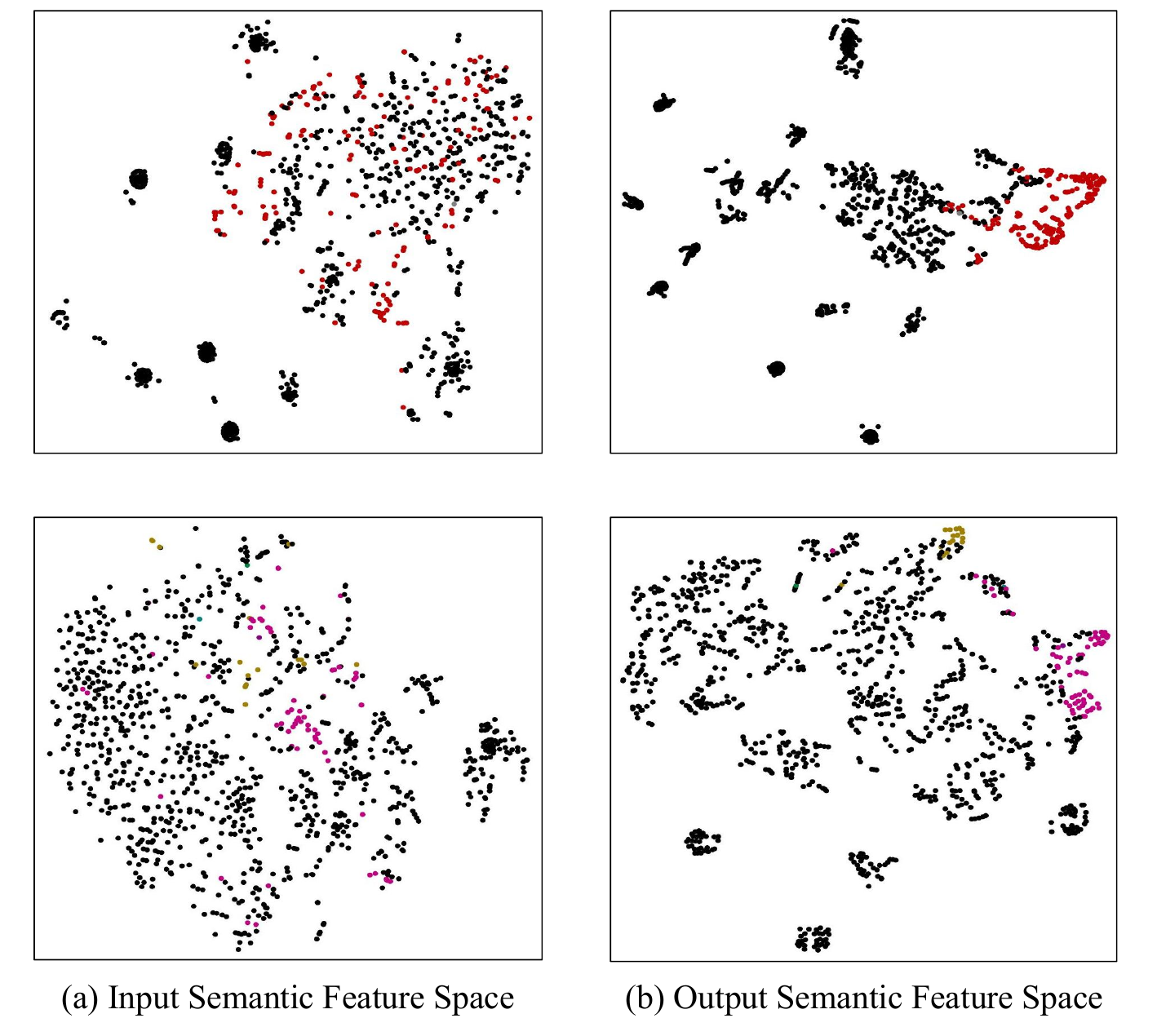}
    \caption{t-SNE visualization results on the semantic feature space.
    The first row illustrates the feature distribution for single-sounding sources, while the second row displays the feature distribution for multi-sounding sources.}
\vspace{-1mm}
    \label{fig:t-sne visualization}
\end{figure}
\subsection{Feature Visualization}
We enhance the interpretability of our model through the visualization of feature maps. Fig.~\ref{fig:feature map visualizaiton} presents the feature maps extracted from the visual backbone and those processed by the Cross-modal FPN across various frames. As depicted in Fig.~\ref{fig:feature map visualizaiton} (c) and (d), we apply principal component analysis (PCA)~\cite{shlens2014tutorial} to extract the principal feature channels and overlay them on the original images. The highlighted regions indicate the areas of focus within the model. Additionally, we reduce the dimensionality of the feature channels to 16 and visualize each channel separately, as shown in Fig.~\ref{fig:feature map visualizaiton} (e) and (f). 
It is evident that feature maps, when not integrated with audio features, predominantly highlight background regions. Conversely, when audio features are incorporated, the model accurately focuses on sound sources, thereby enhancing the contours of targets. This is demonstrated by the instance of the man with the accordion in the upper example and the woman with the guitar and piano in the lower example, where the incorporation of audio features renders them more distinct.

In Fig.~\ref{fig:t-sne visualization}, we apply t-SNE visualization~\cite{van2008visualizing} on the AVSS to further illustrate that our model could facilitate the clustering of semantic regions. The color coding in our visual representations corresponds to different semantic categories. Initially, feature points derived from both single-source and multi-source contexts are intermingled, reflecting a hybrid fusion of semantic content. Our method, CCFormer, effectively disentangles these mixed points, thereby facilitating the clustering of foreground points that share similar semantics. This capability allows for a clear distinction between foreground and background points.

\section{Future Work Directions}
The rapid development of Multimodal Large Language Models (MLLMs)\cite{lai2024lisa,wang2024llm} and unified models\cite{li2024omg,zhang2024psalm} has garnered significant interest in the research community. In our future work, we plan to extend CCFormer to incorporate additional modalities, such as natural language, while maintaining its core concept. This expansion aims to facilitate seamless integration of multiple cross-modal video segmentation tasks. Furthermore, we see potential in combining Large Language Models (LLMs)~\cite{touvron2023llama,liu2024visual} with our proposed modules. By leveraging the strong semantic understanding of LLMs alongside CCFormer’s temporal modeling capabilities, we aim to enhance the model’s performance in video reasoning and segmentation tasks.

\section{Conclusion}
In this paper, we propose CCFormer, a Complementary and Contrastive Transformer to address the challenges associated with insufficient temporal correlation modeling and cross-modal fusion in audio-visual segmentation. The Multi-query Transformer Module is introduced to perform the complementary and comprehensive combination of frame-level and temporal-level information. We further propose the Early Integration Module and Bi-modal Contrastive Loss to effectively integrate and align the multimodal features. Extensive experiments conducted on the AVSBench-object and AVSBench-semantic benchmarks demonstrate that our approach consistently achieves state-of-the-art results. We believe that the insights provided by our work will contribute significantly to ongoing and future research in multimodal learning.

\bibliographystyle{IEEEtran}
\bibliography{ieeetran}

\vfill

\end{document}